\documentclass[11pt]{article}
\usepackage{acl2016}
\usepackage{times}

\usepackage{url}
\usepackage{latexsym}
\usepackage{booktabs}
\usepackage{multirow}
\usepackage{graphicx}
\usepackage{paralist}
\usepackage{mathtools}
\usepackage{dingbat}
\usepackage{subcaption}
\usepackage{balance}
\usepackage{gensymb}
\usepackage{marginnote}
\usepackage{adjustbox}
\usepackage{lingmacros}

\aclfinalcopy 
\def\aclpaperid{19} 

\usepackage{color}
\newcommand{\todo}[1]{}
\renewcommand{\todo}[1]{{\color{red} TODO: {#1}}}

\title{A Proposal for Linguistic Similarity Datasets Based on Commonality Lists}

\author{Dmitrijs Milajevs \\
  Queen Mary University of London \\
  London, UK \\
  {\tt d.milajevs@qmul.ac.uk} \\\And
  Sascha Griffiths \\
  Queen Mary University of London \\
  London, UK \\
  {\tt s.griffiths@qmul.ac.uk} \\
}

\date{}

\begin{document}

\maketitle

\begin{abstract}
Similarity is a core notion that is used in psychology and two branches of linguistics: theoretical and computational. The similarity datasets that come from the two fields differ in design: psychological datasets are focused around a certain topic such as fruit names, while linguistic datasets contain words from various categories. The later makes humans assign low similarity scores to the words that have nothing in common and to the words that have contrast in meaning, making similarity scores ambiguous. In this work we discuss the similarity collection procedure for a multi-category dataset that avoids score ambiguity and suggest changes to the evaluation procedure to reflect the insights of psychological literature for word, phrase and sentence similarity. We suggest to ask humans to provide a list of commonalities and differences instead of numerical similarity scores and employ the structure of human judgements beyond pairwise similarity for model evaluation. We believe that the proposed approach will give rise to datasets that test meaning representation models more thoroughly with respect to the human treatment of similarity.
\end{abstract}

\section{Introduction}
\label{sec:introduction}

Similarity is the degree of resemblance between two objects or events \cite{WCS:WCS1282} and plays a crucial role in psychological theories of knowledge and behaviour, where it is used to explain such phenomena as classification and conceptualisation. \textit{Fruit} is a \emph{category} because it is a practical generalisation. Fruits are sweet and constitute deserts, so when one is presented with an unknown fruit, one can hypothesise that it is served toward the end of a dinner.

Generalisations are extremely powerful in describing a language as well. The verb \textit{runs} requires its subject to be singular. \textit{Verb}, \textit{subject} and \textit{singular} are categories that are used to describe English grammar. When one encounters an unknown word and is told that it is a verb, one will immediately have an idea about how to use it assuming that it is used similarly to other English verbs.

The semantic formalisation of similarity is based on two ideas. The occurrence pattern of a word \emph{defines} its meaning \cite{firth1957lingtheory}, while the difference in occurrence between two words \emph{quantifies} the difference in their meaning \cite{harris1954distributional}. From a computational perspective, this motivates and guides development of similarity components that are embedded into natural language processing systems that deal with tasks such as
word sense disambiguation \cite{Schutze:1998:AWS:972719.972724},
information retrieval \cite{Salton:1975:VSM:361219.361220,Milajevs:2015:IMN:2808194.2809448},
machine translation \cite{Dagan:1993:CWS:981574.981596},
dependency parsing \cite{hermann-blunsom:2013:ACL2013,andreas-klein:2014:P14-2},
and dialogue act tagging \cite{kalchbrenner-blunsom:2013:CVSC,milajevs-purver:2014:CVSC}.

Because it is difficult to measure performance of a single (similarity) component in a pipeline, datasets that focus on similarity are popular among computational linguists. Apart from a pragmatic attempt to alleviate the problems of evaluating similarity components, these datasets serve as an empirical test of the hypotheses of Firth and Harris, bringing together our understanding of human mind, language and technology.

Two datasets, namely MEN \cite{Bruni:2012:DST:2390524.2390544} and SimLex-999 \cite{hill2014simlex}, are currently widely used. They are designed especially for meaning representation evaluation and surpass datasets stemming from psychology \cite{1986-13502-00119860101}, information retrieval \cite{2002:PSC:503104.503110} and computational linguistics \cite{Rubenstein:1965:CCS:365628.365657} in quantity by having more entries and, in case of SimLex-999, attention to the evaluated relation by distinguishing similarity from relatedness. The datasets provide similarity (relatedness) scores between word pairs.

In contrast to linguistic datasets which contain randomly paired words from a broad selection, datasets that come from psychology contain entries that belong to a single category such as \textit{verbs of judging} \cite{FILLENBAUM197454} or \textit{animal terms} \cite{HENLEY1969176}. The reason for category oriented similarity studies is that ``stimuli can only be compared in so far as they have already been categorised as identical, alike, or equivalent at some higher level of abstraction'' \cite{turner1987rediscovering}. Moreover, because of the \emph{extension effect} \cite{medin1993respects}, the similarity of two entries in a context is less than the similarity between the same entries when the context is extended. ``For example, \textit{black} and \textit{white} received a similarity rating of 2.2 when presented by themselves; this rating increased to 4.0 when \textit{black} was simultaneously compared with \textit{white} and \textit{red} (\textit{red} only increased 4.2 to 4.9)'' \cite{medin1993respects}. In the first case \textit{black} and \textit{white} are more dissimilar because they are located on the extremes of the greyscale, but in the presence of \textit{red} they become more similar because they are both monochromes.

Both MEN and SimLex-999 provide pairs that do not share any similarity to control for false positives, and they do not control for the comparison scale. This makes similarity judgements ambiguous as it is not clear what low similarity values mean: incomparable notions, contrast in meaning or even the difference in comparison context. SimLex-999 assigns low similarity scores to the incomparable pairs (0.48, \textit{trick} and \textit{size}) and to antonymy (0.55, \textit{smart} and \textit{dumb}), but \textit{smart} and \textit{dumb} have relatively much more in common than \textit{trick} and \textit{size}!

The present contribution investigates how a similarity dataset with multiple categories should be built and considers what sentence similarity means in this context.

\section{Dataset Construction}

\paragraph{Human similarity judgements}

To build a similarity dataset that contains non-overlapping categories, one needs to avoid comparison of incomparable pairs. However, that itself requires an a priori knowledge of item similarity or belongingness to a category, making the problem circular. To get out of this vicious circle, one might erroneously  refer to an already existing taxonomy such as WordNet \cite{Miller:1995:WLD:219717.219748}. But in case of similarity, as \newcite{turney2012domain} points out, categories that emerge from similarity judgements are different from taxonomies. For example, \textit{traffic} and \textit{water} might be considered to be similar because of a functional similarity exploited in hydrodynamic models of traffic, but their lowest common ancestor in WordNet is \textit{entity}.

Since there is no way of deciding upfront whether there is a similarity relation between two words, the data collection procedure needs to test for both: relation existence and its strength. Numerical values, as has been shown in the introduction, do not fit this role due to ambiguity. One way to avoid the issue is to avoid asking humans for numerical similarity judgements, but instead to ask them to list commonalities and differences between the objects. As one might expect, similarity scores correlate with the number of listed commonalities \cite{markman1991commonalities,Markman1996,medin1993respects}. For incomparable pairs, the commonality list should be empty, but the differences will enumerate properties that belong to one entity, but not to another \cite{markman1991commonalities,medin1993respects}. 

Verbally produced features (norms) for empirically derived conceptual representation of \newcite{McRae2005} is a good example of what and how the data should be collected. But in contrast to \newcite{McRae2005}---where explicit comparison of concepts was avoided---participants should be asked to produce commonalities as part of similarity comparison.

\paragraph{The entries in the dataset}

So far, we have proposed a similarity judgement collection method that is robust to incomparable pairings. It also naturally gives rise to categories, because the absence of a relation between two entries means the absence of a common category. It still needs to be decided which words to include in the dataset.

To get a list of words that constitute the dataset, one might think of categories such as \textit{sports}, \textit{fruits}, \textit{vegetables}, \textit{judging verbs}, \textit{countries}, \textit{colours} and so on. Note, that at this point its acceptable to think of categories, because later the arbitrary category assignments will be reevaluated. Once the list of categories is ready, each of them is populated with category instances, e.g.~\textit{plum}, \textit{banana} and \textit{lemon} are all \textit{fruits}.

When the data is prepared, humans are asked to provide commonalities and differences between all pairs of every group. First, all expected similarities are judged, producing a dataset that can be seen as a merged version of category specific datasets. At this point, a good similarity model should provide meaning representation that are easily split to clusters: \textit{fruit} members and \textit{sport} members have to be separable.

Intra-category comparisons should be also performed, but because it is impractical to collect all possible pairwise judgements between the number of words of magnitude of hundreds, a reasonable sample should be taken. The intra-category comparisons will lead to unexpected category pairings, such as \textit{food} that contains \textit{vegetables} and \textit{fruits}, so the sampling procedure might be directed by the discovery of comparable pairs: when a \textit{banana} and \textit{potato} are said to be similar, \textit{fruits} and \textit{vegetables} members should be more likely to be assessed.

Given the dynamic nature of score collection, we suggest setting up \emph{a game with a purpose} (see \newcite{VenhuizenBasileEvangBos2013IWCS} an example) where players are rewarded for contributing their commonality lists. Another option would be to crowdsource the human judgements \cite{doi:10.1080/17470218.2015.1051065}.

\paragraph{Evaluation beyond proximity}

Human judgements validate the initial category assignment of items and provide new ones. If a category contains a superordinate, similarity judgements arrange category members around it \cite{1986-13502-00119860101}. For example, similarity judgements given by humans arrange fruit names around the word \textit{fruit} in such a way that it is their nearest neighbour, making \textit{fruit} the \emph{focal point} of the category of \textit{fruits}.

As an additional evaluation method, the model should be able to retrieve focal points. Therefore, a precaution should be taken before human judgement collection. If possible, categories should contain a superordinate.

Similarity evaluation needs to focus on how well a model is able to recover human similarity intuitions expressed as groupings, possibly around their focal points. We propose to treat it as a soft multi-class clustering problem \cite{White:2015:WSE:2838931.2838932}, where two entities belong to the same class if there is a similarity judgement for them (e.g.~\textit{apple} and \textit{banana} are similar because they are \textit{fruits}) and the strength is proportional to the number of such judgements, so we could express that \textit{apple} is more a \textit{fruit} than it is a \textit{company}.

In contrast to the current evaluation based on correlation, models also need to be tested on the geometric arrangement of subordinates around the focal points, as only the proximity based evaluation does not capture this \cite{1986-13502-00119860101}.

\section{Sentence Similarity}

The question of sentence similarity is more complex because sentences in many ways are different entities than words. Or are they? Linguistics has recently often pointed toward a continuum which exists between words and sentences \cite{jackendoff2012cambridge}. \newcite{jackendoff2005nature}, for example, point out that there is good evidence that ``human memory must store linguistic expressions of all sizes.'' These linguistic expressions of variable size are often called \emph{constructions}. Several computational approaches to constructions have been proposed \cite{Gaspers2011,chang2012computational}, but to the authors' best knowledge they do not yet feature prominently in natural language processing.

To be able to measure the similarity of phrases and sentences in the proposed framework, we need to be able to identify what could serve as commonalities between them. So what are they? First of all, words, sentences and other constructions draw attention to \emph{states of affairs} around us. Also, sentences are similar to others with respect to the functions they perform \cite[p. 288]{Winograd1983}. 

\paragraph{Prototype effects}

As \newcite{tomasello2009cultural} points out, speakers of English can make sense of phrases like \textit{X floosed Y the Z} and \textit{X was floosed by Y}. This is due to their similarity to sentences such as \textit{John gave Mary the book} and \textit{Mary was kissed by John} respectively. Thus, \textit{X floosed Y the Z} is clearly a \emph{transfer of possession} or \emph{dative} \cite{bresnan2007predicting}.

The amount in which sentences are similar, at least to a certain extent, corresponds to the function of a given sentence (the ideational function \cite[p. 288]{Winograd1983} especially). \newcite{tomasello1998return} points out that sentence-level constructions show \emph{prototype effects} similar to those discussed above for lexical systems (e.g.~colours). Consider the following sentences:
\begin{compactitem}
    \item \textit{John gave Mary the book.} is a example of an \emph{Agent Causes Transfer} construction. These usually are build around words such as \textit{give, pass, hand, toss, bring, etc.}
      \item \textit{John promised Mary the book.} is a example of an \emph{Conditional transfer} construction. These usually are build around words such as \textit{promise, guarantee, owe, etc.}
\end{compactitem}

As soon as one has such a prototype network, one can actually decide sentence similarity as one can say with respect to what prototypes sentences and utterances are similar. In this case, a common sentence prototype serves the same role as commonality between words.

\paragraph{Similarity in context}

However, prototype categories work on the semantic-grammatical level, and might be handled by \emph{similarity in context}: a noun phrase can be similar to a noun as in \textit{female lion} and \textit{lioness}, and to another noun phrases as in \textit{yellow car} and \textit{cheap taxi}. The same similarity principle can be applied to phrases as to words. In this case, similarity is measured in context, but it is still a comparison of the phrases' head words of which meaning is modified by arguments they appear with \cite{Kintsch2001173,mitchell-lapata:2008:ACLMain,mitchell2010composition,Dinu:2010:MDS:1870658.1870771,Baroni2010nouns,thater-furstenau-pinkal:2011:IJCNLP-2011,Seaghdha:2011:PMS:2145432.2145545}. With verbs this idea can be applied to compare transitive verbs with intransitive. For example, \textit{to cycle} is similar to \textit{to ride a bicycle}.

Sentential similarity might be treated as the similarity of the heads in the contexts. That is, the similarity between \textit{sees} and \textit{notices} in \textit{John \textbf{sees} Mary} and \textit{John \textbf{notices} a woman}. This approach abstracts away from grammatical differences between the sentences and concentrates on semantics and fits the proposed model as the respect for the head, which is a lexical entity, has to be found \cite{corbett1993heads}.


\paragraph{Attention attraction}

But still, what about pragmatics? As \newcite{Steels2008} points out, sentences and words direct attention and do not always directly point or refer to entities and actions in the world. For example, he points to the fact that if a person asks another person to \textit{pass the wine} they are actually asking for the \textit{bottle}. The speaker just attracts attention to an object of perception in a given situation.

\paragraph{Grammaticalisaton and lexicalisaton}

There are several ways in which a sentence can both be \emph{grammaticalised} and \emph{lexicalised}. For example, \textit{No} and \textit{I've seen John eating them} are similar sentences because they lexicalise the same answer to the question \textit{Do we have cookies?} More generally, this gives rise to dialogue act tags: for another way of utterance categorisation, refer to the work of \newcite{kalchbrenner-blunsom:2013:CVSC} and \newcite{milajevs-purver:2014:CVSC}.

Thus, questions which the sentences answer, are valid respects for similarity explanation, as well as entailment, paraphrase \cite{White:2015:WSE:2838931.2838932} or spatial categories \cite{ritter-EtAl:2015:*SEM2015}. This also motivates the approach of treating sentences on their own and encoding the meaning of a sentence into a vector in such a way that similar sentences are clustered together \cite{DBLP:journals/corr/abs-1003-4394,baroni2014frege,Socher:2012:SCT:2390948.2391084,wieting2015paraphrase,DBLP:journals/corr/HillCK16}.

\paragraph{Discourse fit}

If one conceptualises sentence similarity with respect to a discourse, then one might ask how different sentences fit in to such a discourse. \newcite{Griffiths2015a} tried to construct two versions of the same dialogue using a bottom-up method. They deconstructed a certain dialogue in a given domain---a receptionist scenario---into \textit{greetings}, \textit{directions} and \textit{farewells}. They used a small custom made corpus for this purpose and created the two dialogues by having people rate the individual utterances by friendliness. The resulting two dialogues were surprisingly uneven. The dialogue was supposed to give instructions to a certain location within a building. The ``friendly version'' was very elaborated and consisted of several sentences:
\enumsentence{
The questionnaire is located in room
Q2-102. That is on the second floor. If
you turn to your right and walk down
the hallway. At the end of the floor you
will find the stairs. Just walk up the
stairs to the top floor and go through
the fire door. The room is then straight
ahead.}
The sentence which served the same purpose in the ``neutral version'' was a fairly simple sentence:
\enumsentence{
The questionnaire is located in Q2-102.
}
Often the same function of a given sentence in a dialogue can be performed by as little as one word or several phrases or a different sentence or even a complete story. 

\paragraph{Language sub-systems and strategies}

\newcite{Steels2010a} introduces the idea of language sub-systems and language strategies. A language subsystem are the means of expressing certain related or similar meanings. Examples of such sub-systems include:
\begin{compactitem}
    \item Lexical systems which express colours.
    \item Morphological devices to encode tenses.
    \item Usage of word order to express relations between agent and patient.
\end{compactitem}  
The later is an illustration of a language strategy. In English agent-patient relations are mainly encoded by syntax whereas German would use intonation and a combination or articles and case to convey the same information. Russian, in contrast, will use morphological devices for the same purpose. Hence, for some purposes the entities which are similar may not be of clearly delineated categories such as ``word'' or ``sentence'' but may be of chunks of language which belong to the same sub-system. 

Above we identified seven criteria by which sentence similarity can be compared. The instructions for the sentence similarity judgement tasks may incorporate the criteria as hints for human participants during data collection.

\section{Conclusion}

In this contribution we discussed the notion of similarity from an interdisciplinary perspective. We contrasted properties of the similarity relation described in the field of psychology with the characteristics of similarity datasets used in computational linguistics. This lead to the recommendations on how to improve the later by removing low score ambiguity in a multi-category similarity dataset.

In the future, a multi-category similarity dataset should be build that allow evaluation of vector space models of meaning by not only measuring proximity between the points, but also their arrangement with respect to clusters. The same ideas can be used to build phrase- and sentence-level datasets. However, we leave the exact sentence similarity criteria selection for future work in this area.

On a broader perspective, this work highlights psychological phenomena that being incorporated into the models of meaning are expected to improve their performance.

\section*{Acknowledgements}

We thank the anonymous reviewers for their comments. Support from EPSRC grant EP/J002607/1 is gratefully acknowledged by Dmitrijs Milajevs.
Sascha Griffiths is supported by ConCreTe: the project ConCreTe acknowledges the financial support of the Future and Emerging Technologies (FET) programme within the Seventh Framework Programme for Research of the European Commission, under FET grant number 611733.

\balance
\bibliographystyle{acl2016}
\bibliography{references,dmilajevs_publications}

\end{document}